\documentclass[11pt]{article}

\usepackage[preprint]{acl}

\usepackage{times}
\usepackage{latexsym}
\usepackage[T1]{fontenc}
\usepackage[utf8]{inputenc}
\usepackage{microtype}
\usepackage{inconsolata}
\usepackage{graphicx}
\usepackage{booktabs}
\usepackage{url}
\usepackage{tabularx}
\usepackage{listings}
\lstdefinestyle{tinyschiller}{
  language=Python,
  basicstyle=\ttfamily\footnotesize,
  keywordstyle=\bfseries,
  commentstyle=\itshape\color[gray]{0.4},
  stringstyle=\ttfamily,
  showstringspaces=false,
  breaklines=true,
  columns=fullflexible,
  xleftmargin=0pt,
  aboveskip=6pt,
  belowskip=2pt,
}
\newcommand{\code}[1]{\texttt{#1}}
\newcommand{\field}[1]{\textsc{#1}}

\title{\textsc{tiny\_schiller:} A Drop-In German Drama Corpus for \\ Small Language Models}

\author{Mark Schutera \\
  Duale Hochschule Baden-W\"urttemberg Ravensburg \\
  \texttt{schutera@dhbw-ravensburg.de} \\}

\begin{document}
\maketitle

\begin{abstract}
\textsc{tiny\_schiller} closes the small-language-model prototyping, fine-tuning, education, 
and research gap for German literary text, providing a single-file, drop-in counterpart to 
Karpathy's \textsc{tiny\_shakespeare}. The available German literary corpora are larger and richer,
 but require parser engineering before a single line of training or fine-tuning code can run. 
 \textsc{tiny\_schiller} is a 2.07-megabyte single file of eleven public-domain Schiller dramas, 
 sourced from DraCor's GerDraCor export~(CC0) and processed by deterministic parser engineering. Character-level, 
 GPT-2 byte-pair encoding, and \code{cl100k\_base} tokenization splits, an instruction-formatted 
 dialogue-completion split, and 89 per-character persona splits load from a single HuggingFace call.
 A small language model literally reaches German literary text in one line of code.
\end{abstract}

\section{Introduction}
\label{sec:intro}

\paragraph{The German tiny-corpus gap.} Small literary corpora are a fixture of small-scale
language modelling. \textsc{tiny\_shakespeare}~\citep{karpathy2015charrnn}, popularised
through \textsc{nanoGPT}~\citep{karpathy2022nanogpt}, lets a practitioner train a working
transformer from scratch in minutes on a single laptop, and serves equally well as a
fine-tuning target on a stylistically distinctive register. The BabyLM Challenge~\citep{warstadt2023babylm} has since established small fixed corpora as a first-class unit
of LM research. The artifact behind \textsc{tiny\_shakespeare}'s reach is a single 1.1~MB
file. The pipeline assumed by most educational and research material reads it directly.

No comparable single-file, cleaned, German counterpart exists. Practitioners who want to
prototype, fine-tune, or use a non-English literary register for education and research
currently fall back either on multi-gigabyte web crawls or on ad-hoc text downloads from
sources such as Projekt Gutenberg-DE~\citep{projektgutenberg}, without provenance,
without documented cleaning, and without reproducible preprocessing.
Larger and richer German corpora exist: The Deutsches Textarchiv~\citep{deutschestextarchiv}
and DraCor~\citep{fischer2019dracor, borner2023dockerizing} are deeper resources by every
metric except one.\\

\paragraph{The Metric is Friction.} If the German literary corpora are deeper,
better-annotated, and longer-established than \textsc{tiny\_shakespeare}, why has no German
equivalent of \textsc{tiny\_shakespeare}-scale small-LM emerged?
Not because the data is missing. Because reaching the data costs an attention budget that
small-scale prototyping, education, and research rather would not pay. For German, there is no single cleaned file that drops into 
a reproducible small-LM pipeline without parser engineering.
The available German corpus resources require non-trivial preliminaries before a single token
reaches a model: TEI/XML parsers, namespace handling, edition normalisation, speaker-tag
reconciliation, encoding fixes for legacy umlauts and quote glyphs, and, in DraCor's case,
a container stack to reproduce the corpus build at all~\citep{borner2023dockerizing}.
The work is not technically difficult. However, it is \emph{upstream} of the modelling task and
competes with it for a finite attention budget. In an educational setting it consumes the
first session before any model has seen any text. In a prototyping or research setting it
dominates the first day before any tokenizer has been compared. This paper refers to this aggregate
of small, individually tractable preprocessing steps as \emph{single-file friction}: The
gap between ``a corpus exists in some form'' and ``a corpus is ready for the downstream small-LM''.

\textsc{tiny\_shakespeare}'s reach in English-language education and research does not come
from its size. It comes from having no friction. \paragraph{\textsc{tiny\_schiller}} is built to remove
the friction for German, which means contributing four small things together:

\begin{itemize}
    \setlength\itemsep{0.2em}
    \item \textbf{A cleaned, single-file release.} A 2.07~MB UTF-8 file produced by a
    deterministic pipeline, drop-in for small-LM trainers from scratch~(including
    \textsc{nanoGPT}) and for standard fine-tuning pipelines (\S\ref{sec:audit}).
    \item \textbf{Tokenization splits matching standard small-LM workflows.}
    Character-level, GPT-2 BPE, and \code{cl100k\_base} token streams are precomputed
    (\S\ref{sec:preprocessing}).
    \item \textbf{Fine-tuning Parquets.} Whole-work, instruction-formatted
    dialogue-completion (\code{instruct.parquet}), and 89 per-character persona parquets
    are released for supervised fine-tuning (\S\ref{sec:splits}).
    \item \textbf{A reference fine-tune.} End-to-end usability on a single consumer GPU,
    documented with concrete numbers rather than asserted (\S\ref{sec:reference}).
    \item \textbf{An agent-ready data card.} A concise Markdown summary~(\code{DATA\_CARD.md}),
    structured as a data statement~\citep{bender2018datastatements}, that documents provenance
    and intended use and enables quick, robust corpus ingestion by agentic tooling.\footnote{\url{https://github.com/schutera/tiny_schiller}.}
\end{itemize}
% TODO check why availability is section 9?

Beyond friction removal, the BPE and \code{cl100k\_base} splits double as a testbed for
a known second-order effect: English-trained tokenizers encode archaic German poorly, a
gap documented at scale~\citep{rust2021goodbpe, petrov2023unfairness, ahia2023costsame}.
This paper reports the corpus-level fertility number for this specific testbed in
\S\ref{sec:preprocessing}.

% --------------------------------------------------------------------------------------------------------
% --------------------------------------------------------------------------------------------------------

\section{On \textsc{tiny\_schiller}}
\label{sec:corpus}

\textsc{tiny\_schiller} comprises eleven dramatic works by Friedrich Schiller
(Table~\ref{tab:works}), all in the public domain and sourced from DraCor's GerDraCor
plain-text export (CC0).\footnote{\url{https://dracor.org/api/v1/corpora/ger/}} The texts span
Schiller's dramatic career from \emph{Die R\"auber} (1781) to \emph{Wilhelm Tell} (1804) and
include the canonical mature dramas. Poetry, prose fiction, historical writings, and
aesthetic essays are excluded to keep the corpus stylistically homogeneous and dialog-dense.

Schiller is a deliberate \emph{single-author} choice. The eleven dramas are stylistically
homogeneous, populated by a small set of vivid characters whose voices remain consistent
across acts, and exercise the orthographic edges of German (umlauts, the eszett, French
loanword diacritics, and \guillemotleft\,angle quotes\,\guillemotright{}) that modern news
German rarely touches. Goethe and Lessing are reasonable alternative choices; this point is
revisited in \S\ref{sec:use}.

\begin{table}[t]
\centering
\footnotesize
\begin{tabularx}{\columnwidth}{@{}Xr@{}}
\toprule
\textbf{Work} & \textbf{Personas} \\
\midrule
Die R\"auber & 20 \\
Die Verschw\"orung des Fiesco zu Genua & 25 \\
Kabale und Liebe & 10 \\
Don Carlos, Infant von Spanien & 34 \\
Wallensteins Lager & 15 \\
Die Piccolomini & 23 \\
Wallensteins Tod & 24 \\
Maria Stuart & 23 \\
Die Jungfrau von Orleans & 39 \\
Die Braut von Messina (test split) & 11 \\
Wilhelm Tell (test split) & 51 \\
\bottomrule
\end{tabularx}
\caption{Works included in \textsc{tiny\_schiller} and the number of distinct speaker personas (with at least one turn) present in each work.}
\label{tab:works}
\end{table}

% --------------------------------------------------------------------------------------------------------
% --------------------------------------------------------------------------------------------------------

\section{Making the File Drop-In}
\label{sec:audit}

\begin{table}[b]
\centering
\small
\begin{tabularx}{\columnwidth}{@{}l>{\raggedleft\arraybackslash}X@{}}
\toprule
\textbf{\textsc{tiny\_schiller} Statistic} & \textbf{Value} \\
\midrule
\textsc{File} characters              & 2{,}019{,}857 \\
\textsc{File} bytes (UTF-8)           & 2{,}067{,}041 \\
\textsc{File} encoding                & UTF-8 (no BOM) \\
\textsc{File} line endings            & LF \\
\addlinespace[0.35em]
\textsc{Charset} unique codepoints    & 88 \\
\textsc{Charset} \guillemotleft{} / \guillemotright{} & 101 / 101 \\
\addlinespace[0.35em]
\textsc{Drama} works                  & 11 \\
\textsc{Drama} speaker-tag format     & \code{SPEAKER:\textbackslash n} \\
\bottomrule
\end{tabularx}
\caption{\code{tiny\_schiller.parsed.txt} summary statistics: File, character-set, and dramatic-structure
properties.}
\label{tab:audit}
\end{table}

\code{tiny\_schiller.parsed.txt}, the single-file artifact: 2{,}019{,}857 characters
in 2.07~MB of UTF-8 (no BOM), 1.88$\times$ \textsc{tiny\_shakespeare} in bytes but fewer
tokens under BPE (\S\ref{sec:preprocessing}). ``No BOM'' means the file does not begin with a
UTF-8 byte-order mark, which avoids spurious leading characters in downstream tooling. All other
released artifacts are deterministic derivatives.

The file is produced by \code{scripts/parse.py} from the concatenated raw downloads, performing
(a) mechanical normalisation of upstream typesetting artefacts and (b) unification of three
speaker-tag styles into one canonical form. Each step is small in isolation; together they
eliminate the parser-engineering burden described in \S\ref{sec:intro}. 

\iffalse 
From this canonical
speaker-tagged file, the release deterministically derives (c) the tokenization splits
(\S\ref{sec:preprocessing}) and (d) the HuggingFace fine-tuning parquets (\S\ref{sec:splits}).
\fi

\paragraph{Mechanical normalisation.} Convert CRLF to LF; collapse $\geq$3 blank lines to two;
strip the narrow no-break space (NNBSP, U+202F) found in some upstream editions;
and normalise en-dashes. These reversible byte-level edits match the conventions assumed by
tokenizer training scripts and small-LM preparation utilities.

\paragraph{Speaker-tag unification.} Upstream editions mix three styles (inline
\textsc{Speaker.\,text}, standalone \textsc{Speaker.}, and standalone \textsc{Speaker:});
all are rewritten into \code{SPEAKER:\textbackslash ntext}, the
\textsc{tiny\_shakespeare}/\textsc{nanoGPT} convention. The regular speaker boundary
feeds the per-character splits (\S\ref{sec:splits}) and matches common assumptions in
computational drama analysis~\citep{moretti2011network, fischer2019dracor}.

% --------------------------------------------------------------------------------------------------------
% --------------------------------------------------------------------------------------------------------

\section{Three Tokenization Splits}
\label{sec:preprocessing}

This release provides three precomputed token streams for standard small-LM workflows: A character
baseline and two widely used subword encodings. Each split follows a deterministic 90/10
train/validation partition and writes \code{train.bin} / \code{val.bin} in the
\textsc{nanoGPT} convention. Table~\ref{tab:tokenizers} reports vocabulary size and
tokenization efficiency (characters per token) on \code{tiny\_schiller.parsed.txt}.

\paragraph{Default and takeaway.} Unless stated otherwise, this paper uses \code{schiller\_bpe}
for token-based reporting in this paper, since GPT-2 BPE is a widely used and stable
baseline in small-LM tooling. The three splits are a character-level baseline
(\code{schiller\_char}, vocabulary 88), GPT-2 BPE~\citep{radford2019gpt2} via
\code{tiktoken}~\citep{sennrich2016bpe, tiktoken}~(\code{schiller\_bpe}), and \code{cl100k\_base}
~(\code{schiller\_cl100k}). The \code{cl100k\_base} split is included to quantify
context-budget effects: On this corpus it yields 3.14 chars/tok vs.\ 2.36 for GPT-2
BPE, reflecting the known cross-lingual inefficiency of English-trained tokenizers on
German documented at scale elsewhere~\citep{rust2021goodbpe, petrov2023unfairness,
ahia2023costsame}.

\begin{table}[hb]
\centering
\small
\begin{tabularx}{\columnwidth}{@{}Xrr@{}}
\toprule
\textbf{Pipeline} & \textbf{Vocab} & \textbf{chars/tok} \\
\midrule
\code{schiller\_char}    & 88      & 1.00 \\
\code{schiller\_bpe}     & 50{,}257 & 2.36 \\
\code{schiller\_cl100k}  & $\approx$100k & 3.14 \\
\bottomrule
\end{tabularx}
\caption{\code{tiny\_schiller.parsed.txt} tokenizer comparison.}
\label{tab:tokenizers}
\end{table}

% --------------------------------------------------------------------------------------------------------
% --------------------------------------------------------------------------------------------------------

\section{Fine-Tuning Parquets}
\label{sec:splits}

Loading any of the data splits from the HuggingFace Hub takes a single line:

\vspace{0.6em}
\begin{lstlisting}
from datasets import load_dataset

ds = load_dataset("mrkschtr/tiny_schiller", split="train")
\end{lstlisting}
\vspace{0.6em}

\paragraph{Three parquet splits.} Where \S\ref{sec:preprocessing} provides
\emph{token streams} for training from scratch, this section provides
\emph{example datasets} for supervised fine-tuning. The release ships three parquet
splits on the HuggingFace
Hub,\footnote{\url{https://huggingface.co/datasets/mrkschtr/tiny_schiller}.} all derived
deterministically from \code{tiny\_schiller.parsed.txt} by
\code{scripts/build\_instruct.py}: A whole-work split, an instruction-formatted
dialogue-completion split (\code{instruct.parquet}), and 89 per-character persona splits.

\paragraph{Schemas.} The whole-work split exposes a single \field{text} field per row
and holds out \emph{Wilhelm Tell} and \emph{Die Braut von Messina} as the test
partition. \code{instruct.parquet} exposes \field{prompt} and \field{completion} fields,
with \field{work} and \field{character} as filtering metadata. The 89 per-character
files share that schema and contain only the rows whose target speaker matches the named
character.

\paragraph{Construction.} The instruct and persona rows are produced by a single sliding
window over the canonical speaker-tagged file: The prompt fills one of six non-persona
dialogue-continuation templates with three preceding speaker turns from one work, and
the completion is the next speaker's turn, formatted as
\code{NAME:\textbackslash ntext}. The per-character files use the same window with four
character-specific persona templates.

\paragraph{From split to trainer.} The instruct and persona splits load through the same
call by pointing the loader at the appropriate parquet file under \code{data/}. The two
schemas reflect different training framings: The whole-work split is shaped for
next-token training, the instruct and persona splits for prompt/completion SFT. A
one-line map that concatenates the prompt and completion fields into a single text field
unifies them before tokenization.

% --------------------------------------------------------------------------------------------------------
% --------------------------------------------------------------------------------------------------------

\section{A Reference Fine-Tune}
\label{sec:reference}

The reference fine-tune is the proof that the one-screen pitch holds: Data drop-in,
supervised fine-tuning, and a working stylistic checkpoint, all on a single consumer
GPU. The full training script fits on a screen:

\vspace{0.6em}
\begin{lstlisting}
from datasets import load_dataset
from transformers import AutoModelForCausalLM, TrainingArguments
from trl import SFTTrainer

ds = load_dataset("mrkschtr/tiny_schiller", data_files="data/instruct.parquet", split="train")
model = AutoModelForCausalLM.from_pretrained("Qwen/Qwen2.5-0.5B-Instruct")
args = TrainingArguments(output_dir="out", per_device_train_batch_size=4, num_train_epochs=2)
SFTTrainer(model, ds, args).train()
\end{lstlisting}
\vspace{0.6em}

\noindent The reference run is a two-stage SFT. Stage~1 trains on the full \code{instruct.parquet}
to teach the base model the dialogue-continuation register; stage~2 specialises a copy
of that checkpoint on a single per-character file (\code{char\_MOOR.parquet}) to
demonstrate persona adaptation. The two stages exercise the two intended fine-tuning
workflows from \S\ref{sec:splits} (corpus-wide stylistic adaptation and per-character
specialisation) in one pipeline, and reports concrete numbers rather than asserting
feasibility.

Persona adaptation completes in three minutes (Table~\ref{tab:training_params}), cheap
enough to iterate on interactively; the full two-stage run still fits inside a single
session on accessible hardware.

\begin{table}[ht]
\centering
\small
\begin{tabularx}{\columnwidth}{@{}lX@{}}
\toprule
\textbf{Parameter} & \textbf{Value} \\
\midrule
Base model & \code{Qwen/Qwen2.5-0.5B-Instruct} \\
Stage 1 dataset & \code{instruct.parquet}, 7{,}454 train / 153 eval (2\% holdout) \\
Stage 2 dataset & \code{char\_MOOR.parquet}, 127 train / 3 eval \\
Epochs & 1. Stage: 2, 2. Stage: 2 \\
Early stopping & eval\_loss, patience 2, eval every 100 steps (not triggered) \\
Batch size & 4 \\
Context length & 512 tokens \\
Learning rate & $2 \times 10^{-4}$ (Stage 1), $1 \times 10^{-4}$ (Stage 2) \\
Weight decay & 0.05 (Stage 2 only) \\
Hardware & NVIDIA RTX 3060 (12 GB VRAM, bf16 precision) \\
Training time & 3.6 hours total (Stage 1: 3 h 34 min, Stage 2: 3 min) \\
Final eval loss & 0.182 (Stage 1) / 0.044 (Stage 2) \\
Final token accuracy & 0.965 (Stage 1) / 0.989 (Stage 2) \\
\bottomrule
\end{tabularx}
\caption{Reference fine-tune: Parameters, hardware, and final-checkpoint metrics.}
\label{tab:training_params}
\end{table}

% --------------------------------------------------------------------------------------------------------
% --------------------------------------------------------------------------------------------------------

\section{Where This Helps and Where It Doesn't}
\label{sec:use}

Given the artifacts of \S\ref{sec:splits} and the reference run of \S\ref{sec:reference},
\textsc{tiny\_schiller} is intended for drop-in use in small-LM prototyping, producing
coherent in-register continuations from a small open base model. The most direct use
cases include:

\begin{itemize}
    \setlength\itemsep{0.2em}
    \item \textbf{Rapid prototyping} of architectures and tokenization choices in
    low-compute settings.
    \item \textbf{Stylistic fine-tuning} of pretrained German or multilingual models on a
    homogeneous literary register.
    \item \textbf{Per-character persona fine-tuning} using the prebuilt per-character
    splits.
    \item \textbf{Education and research}, including small-scale language modelling
    studies on a non-English literary register.
\end{itemize}

\vspace{1em}
The contribution is operational rather than evaluative: No claim is made about generation
quality. As an illustrative random sample, the Karl-Moor persona fine-tune (0.5B model, \S\ref{sec:reference})
continues the prompt \emph{Freiheit?} as follows:

\begin{quote}
\small
\textbf{Moor}\\
\emph{aufspringend.
Frein! Bist du wahnsinnig? Willst du dich von deinen Bubenjahren hofmeistern?
Komm näher – zögerst du wieder vom Pferd, wo die Ehre ist?}\\[0.3em]
\textbf{Schweizer:} \\
\emph{Gebeut, Hauptmann! was sollen wir tun?} \\[0.3em]
\textbf{Moor:} \\
\emph{Steh auf, Schweizer! und rührt sich der Schaum. Wir müssen durch diese Nacht noch
zurückkommen.} \\[0.3em]
\end{quote}

\noindent The corpus is too small to pretrain a competitive general-purpose language model
and does not constitute a benchmark in the sense of a held-out evaluation suite.
Speaker-tag detection is regex-based rather than grammar-based, so while it normalises
the most common plain-text edition styles, it remains a heuristic; edge-case typography,
such as stage directions formatted as speaker turns, survives into the released file and
should be expected.

\paragraph{Beyond Schiller.} The same toolchain applies to other public-domain drama with
explicit speaker tags, such as Goethe, Lessing, and Kleist, and to German translations of
international authors, enabling future \textsc{tiny\_$\langle$author$\rangle$} releases
for cross-author comparison at the same scale. Minor adaptations extend it to non-dramatic
text and to additional languages.

% Acknowledgements section omitted in review version per ACL policy.

\bibliography{references}

@misc{karpathy2015charrnn,
  author = {Karpathy, Andrej},
  title  = {The Unreasonable Effectiveness of Recurrent Neural Networks},
  year   = {2015},
  note   = {Blog post; introduces \texttt{tiny\_shakespeare}},
  howpublished = {\url{https://karpathy.github.io/2015/05/21/rnn-effectiveness/}},
  urldate = {2026-05-01}
}

@misc{karpathy2022nanogpt,
  author = {Karpathy, Andrej},
  title  = {{nanoGPT}},
  year   = {2022},
  howpublished = {\url{https://github.com/karpathy/nanoGPT}},
  urldate = {2026-05-01}
}

@inproceedings{fischer2019dracor,
  author    = {Fischer, Frank and B\"orner, Ingo and G\"obel, Mathias and Hechtl, Angelika and Kittel, Christopher and Milling, Carsten and Trilcke, Peer},
  title     = {Programmable Corpora: Introducing {DraCor}, an Infrastructure for the Research on European Drama},
  booktitle = {Proceedings of the Digital Humanities Conference ({DH2019})},
  year      = {2019},
  address   = {Utrecht, Netherlands},
  doi       = {10.5281/zenodo.4284002}
}

@inproceedings{borner2023dockerizing,
  author    = {B\"orner, Ingo and Trilcke, Peer and Milling, Carsten and Fischer, Frank and Sluyter-G\"athje, Henny},
  title     = {Dockerizing {DraCor}: A Container-based Approach to Reproducibility in Computational Literary Studies},
  booktitle = {Proceedings of the Digital Humanities Conference ({DH2023})},
  year      = {2023},
  address   = {Graz, Austria},
  doi       = {10.5281/zenodo.8107836}
}

@inproceedings{warstadt2023babylm,
  author    = {Warstadt, Alex and Mueller, Aaron and Choshen, Leshem and Wilcox, Ethan and Zhuang, Chengxu and Ciro, Juan and Mosquera, Rafael and Paranjabe, Bhargavi and Williams, Adina and Linzen, Tal and Cotterell, Ryan},
  title     = {Findings of the {BabyLM} Challenge: Sample-Efficient Pretraining on Developmentally Plausible Corpora},
  booktitle = {Proceedings of the BabyLM Challenge at CoNLL 2023},
  publisher = {Association for Computational Linguistics},
  address   = {Singapore},
  year      = {2023},
  pages     = {1--34},
  doi       = {10.18653/v1/2023.conll-babylm.1}
}

@inproceedings{petrov2023unfairness,
  author    = {Petrov, Aleksandar and La Malfa, Emanuele and Torr, Philip H. S. and Bibi, Adel},
  title     = {Language Model Tokenizers Introduce Unfairness Between Languages},
  booktitle = {Advances in Neural Information Processing Systems ({NeurIPS})},
  year      = {2023},
  url       = {https://proceedings.neurips.cc/paper_files/paper/2023/hash/74bb24dca8334adce292883b4b651eda-Abstract-Conference.html}
}

@inproceedings{ahia2023costsame,
  author    = {Ahia, Orevaoghene and Kumar, Sachin and Gonen, Hila and Kasai, Jungo and Mortensen, David R. and Smith, Noah A. and Tsvetkov, Yulia},
  title     = {Do All Languages Cost the Same? {T}okenization in the Era of Commercial Language Models},
  booktitle = {Proceedings of the 2023 Conference on Empirical Methods in Natural Language Processing ({EMNLP})},
  publisher = {Association for Computational Linguistics},
  year      = {2023},
  pages     = {9904--9923},
  doi       = {10.18653/v1/2023.emnlp-main.614}
}

@misc{projektgutenberg,
  title  = {Projekt Gutenberg-{DE}},
  author = {{Cultural Assets GmbH}},
  year   = {2026},
  howpublished = {\url{https://www.projekt-gutenberg.org/}},
  note   = {formerly operated by Hille and Partner},
  urldate = {2026-07-14}
}

@misc{deutschestextarchiv,
  title  = {Deutsches Textarchiv ({DTA})},
  author = {{Berlin-Brandenburgische Akademie der Wissenschaften}},
  year   = {2007},
  howpublished = {\url{https://www.deutschestextarchiv.de/}},
  note   = {ongoing project},
  urldate = {2026-07-14}
}

@techreport{radford2019gpt2,
  author      = {Radford, Alec and Wu, Jeffrey and Child, Rewon and Luan, David and Amodei, Dario and Sutskever, Ilya},
  title       = {Language Models are Unsupervised Multitask Learners},
  institution = {OpenAI},
  year        = {2019},
  url         = {https://cdn.openai.com/better-language-models/language_models_are_unsupervised_multitask_learners.pdf}
}

@inproceedings{sennrich2016bpe,
  author    = {Sennrich, Rico and Haddow, Barry and Birch, Alexandra},
  title     = {Neural Machine Translation of Rare Words with Subword Units},
  booktitle = {Proceedings of the 54th Annual Meeting of the Association for Computational Linguistics ({ACL})},
  publisher = {Association for Computational Linguistics},
  address   = {Berlin, Germany},
  year      = {2016},
  pages     = {1715--1725},
  doi       = {10.18653/v1/P16-1162}
}

@misc{tiktoken,
  author = {{OpenAI}},
  title  = {tiktoken: A Fast {BPE} Tokeniser for Use with {OpenAI}'s Models},
  year   = {2022},
  howpublished = {\url{https://github.com/openai/tiktoken}},
  urldate = {2026-05-01}
}

@techreport{moretti2011network,
  author      = {Moretti, Franco},
  title       = {Network Theory, Plot Analysis},
  institution = {Stanford Literary Lab},
  type        = {Pamphlet},
  number      = {2},
  year        = {2011},
  month       = may,
  url         = {https://litlab.stanford.edu/LiteraryLabPamphlet2.pdf}
}

@article{bender2018datastatements,
  author  = {Bender, Emily M. and Friedman, Batya},
  title   = {Data Statements for Natural Language Processing: Toward Mitigating System Bias and Enabling Better Science},
  journal = {Transactions of the Association for Computational Linguistics ({TACL})},
  volume  = {6},
  year    = {2018},
  pages   = {587--604},
  doi     = {10.1162/tacl_a_00041}
}

@inproceedings{rust2021goodbpe,
  author    = {Rust, Phillip and Pfeiffer, Jonas and Vuli\'c, Ivan and Ruder, Sebastian and Gurevych, Iryna},
  title     = {How Good is Your Tokenizer? On the Monolingual Performance of Multilingual Language Models},
  booktitle = {Proceedings of the 59th Annual Meeting of the Association for Computational Linguistics and the 11th International Joint Conference on Natural Language Processing ({ACL}-{IJCNLP})},
  publisher = {Association for Computational Linguistics},
  year      = {2021},
  pages     = {3118--3135},
  doi       = {10.18653/v1/2021.acl-long.243}
}

\end{document}